\documentclass{article}

\usepackage[preprint]{neurips_2025}

\usepackage{hyphenat}
\usepackage[utf8]{inputenc} 
\usepackage[T1]{fontenc}    
\usepackage{hyperref}       
\usepackage{url}            
\usepackage{booktabs}       
\usepackage{amsfonts}       
\usepackage{nicefrac}       
\usepackage{microtype}      
\usepackage{xcolor}         

\usepackage[utf8]{inputenc} 
\usepackage[T1]{fontenc}    
\usepackage{hyperref}       
\usepackage{url}            
\usepackage{booktabs}       
\usepackage{amsfonts}       
\usepackage{nicefrac}       
\usepackage{microtype}      
\usepackage{xcolor}         

\usepackage{hyperref}
\usepackage{footnote}
\usepackage{caption}
\usepackage{tikz}

\usepackage{mathrsfs}

\usepackage{graphicx}
\usepackage{wrapfig}
\usepackage{caption}
\usepackage{lipsum} 

\usepackage{algpseudocode}
\usepackage[ruled,linesnumbered]{algorithm2e}
\usepackage{amsmath} 
\usepackage{array}
\usepackage{tabularx}
\usepackage{multirow}
\usepackage[most]{tcolorbox}
\usepackage{makecell}
\usepackage{colortbl}
\usepackage{balance}
\usepackage{float}
\usepackage{graphicx}
\usepackage{subcaption}
\usepackage{diagbox}

\usepackage[ruled,linesnumbered]{algorithm2e}

\newcommand{\Rmnum}[1]{\expandafter\@slowromancap\romannumeral #1@}

\newcommand{\ie}{\emph{i.e.},\xspace}

\DeclareMathAlphabet\mathbfcal{OMS}{cmsy}{b}{n}

\newcommand{\eat}[1]{}
\ifodd 1

\newcommand{\TODO}[1]{{\color{red}TODO:{#1}}}

\else

\fi

\usepackage{pifont}%
\tcbset{
  aibox/.style={
    width=400.18663pt,
    top=10pt,
    colback=black!05,
    colframe=black!20,
    colbacktitle=black!25!gray,
    enhanced,
    center,
    attach boxed title to top center={yshift=-0.1in},
    boxed title style={boxrule=0pt,colframe=white,},
  }
}

\tcbset{
  aiboxnotitle/.style={
    width=400.18663pt,
    top=10pt,
    colback=black!05,
    colframe=black!20,
    enhanced,
    center,
  }
}

\tcbset{
  aiboxbreakable/.style={
    width=400.18663pt,
    top=10pt,
    colback=black!05,
    colframe=black!20,
    colbacktitle=black!50,
    enhanced,
    center,
    breakable,
    attach boxed title to top left={yshift=-0.1in,xshift=0.15in},
    boxed title style={boxrule=0pt,colframe=white,},
  }
}
\newtcolorbox{AIBox}[2][]{aibox,title=#2,#1}
\newtcolorbox{AIBoxNoTitle}[1][]{aiboxnotitle}
\newtcolorbox{AIBoxBreak}[2][]{aiboxbreakable,title=#2,#1}
\usepackage{alltt}
\usepackage{listings}
\title{Foundation Models for Scientific Discovery: From Paradigm Enhancement to Paradigm Transition}

\renewcommand\footnotemark{}
\author{
Fan Liu$^{1}$, Jindong Han$^{2}$, Tengfei Lyu$^{1}$, Weijia Zhang$^{1}$, Zhe-Rui Yang$^{1}$,\\ \textbf{Lu Dai}$^{2,1}$, \textbf{Cancheng Liu}$^{1}$, \textbf{Hao Liu}$^{1,2}$\thanks{Correspondence to Hao Liu.} \\
$^1$The Hong Kong University of Science and Technology (Guangzhou), Guangzhou, China \\
$^2$The Hong Kong University of Science and Technology, Hong Kong SAR, China \\
\footnotesize{\texttt{fliu236@connect.hkust-gz.edu.cn; hanjindong01@gmail.com;}}\\
\footnotesize{\texttt{tlyu077@connect.hkust-gz.edu.cn; vegazhang3@gmail.com;}}\\
\footnotesize{\texttt{ldaiae@connect.ust.hk; liuh@ust.hk}}
}

\begin{document}

\maketitle

\vspace{-0.1in}
\begin{abstract}
Foundation models (FMs), such as GPT-4 and AlphaFold, are reshaping the landscape of scientific research. Beyond accelerating tasks such as hypothesis generation, experimental design, and result interpretation, they prompt a more fundamental question: Are FMs merely enhancing existing scientific methodologies, or are they redefining the way science is conducted? In this paper, we argue that FMs are catalyzing a transition toward a new scientific paradigm. We introduce a three-stage framework to describe this evolution: (1) Meta-Scientific Integration, where FMs enhance workflows within traditional paradigms; (2) Hybrid Human-AI Co-Creation, where FMs become active collaborators in problem formulation, reasoning, and discovery; and (3) Autonomous Scientific Discovery, where FMs operate as independent agents capable of generating new scientific knowledge with minimal human intervention. Through this lens, we review current applications and emerging capabilities of FMs across existing scientific paradigms. We further identify risks and future directions for FM-enabled scientific discovery. This position paper aims to support the scientific community in understanding the transformative role of FMs and to foster reflection on the future of scientific discovery. Our project is available at \url{https://github.com/usail-hkust/Awesome-Foundation-Models-for-Scientific-Discovery}. \label{abs}
\vspace{-0.1in}

\end{abstract}

\section{Introduction}

Scientific discovery has historically progressed through a series of methodological paradigms, each redefining how researchers observe, explain, and model the natural world. From the empirical investigations of Galileo and Boyle to the formal theories of Newton and Einstein, science has advanced by transforming observations into abstract and systematic knowledge. Later, computational simulation enabled the exploration of systems too complex for direct experimentation, while the rise of data-driven science in the 21st century reframed discovery as the extraction of statistical patterns from massive datasets. Together, these four paradigms, \ie experimental, theoretical, computational, and data-driven, constitute the foundations of modern scientific practice~\cite{shapin2018scientific,winsberg2019science,breiman2001statistical}.

However, as science increasingly engages with phenomena characterized by emergent behavior, open-endedness, and irreducible complexity, the limitations of existing paradigms have become more evident~\cite{jumper2021highly, nguyen2023climax, romera2024funsearch}. Challenges such as understanding consciousness, modeling protein folding pathways, and predicting social polarization defy reductionist modeling and remain intractable, even in the face of recent advances in machine learning~\cite{more2022interface,chen2022crom}. In fields like drug discovery and material design, the combinatorial explosion of candidate spaces makes exhaustive search infeasible~\cite{lyu2019ultra}. Meanwhile, the rapid accumulation of experimental and observational data has outpaced our capacity to synthesize unifying theories or explanatory frameworks, widening the gap between empirical richness and conceptual understanding. Even state-of-the-art computational models often rely on simplifying assumptions such as linearity, stationarity, or equilibrium, which are fundamentally misaligned with the dynamic, non-linear, and adaptive nature of many real-world systems~\cite{frigg2006models, pearl2009causality}. These tensions underscore a growing mismatch between the increasing complexity of scientific problems and the methodological frameworks currently available to address them.

Foundation Models (FMs)~\cite{bommasani2021opportunities} offer a promising response to these challenges. As large-scale neural networks trained on diverse and extensive datasets, FMs exhibit remarkable adaptability, performing a wide range of tasks via prompting or fine-tuning. Models such as GPT-4~\cite{, DBLP:journals/corr/abs-2410-21276}, AlphaFold~\cite{jumper2021highly}, and DeepSeek~\cite{DBLP:journals/corr/abs-2412-19437} have demonstrated unprecedented capabilities in language understanding, code generation, and scientific reasoning. For instance, AlphaFold~\cite{jumper2021highly} resolved the long-standing protein folding challenge by navigating an intractable configuration space using learned structural priors. FunSearch~\cite{romera2024mathematical}, developed by DeepMind, goes even further, showing that FMs can autonomously propose and validate new mathematical conjectures, rivaling expert-designed algorithms on NP-hard problems. These advances reflect a broader trend: FMs not only accelerate existing scientific workflows, but also begin to change how knowledge is generated, organized, and applied. Unlike previous AI systems that were built for specific tasks, FMs offer a unified architecture capable of handling text, code, and even multi-modal inputs. More importantly, they support new ways of thinking, enabling reasoning, abstraction, and exploration at scale. In this sense, FMs blur the boundary between tool and collaborator, between algorithmic processing and cognitive engagement. This brings us to a critical question: Are foundation models simply enhancing the current scientific paradigm, or are they catalyzing the emergence of a new one?

Throughout history, paradigm shifts in science have not only introduced new tools but also transformed the way science is understood and practiced. Transitions from observation to explanation, or from simulation to data-driven inference, have introduced new epistemologies for formulating problems, generating evidence, and establishing scientific validity. Today, FMs may represent a similar inflection point. By unifying language, code, and multimodal inputs within a single framework, FMs can retrieve literature, formulate hypotheses, simulate complex phenomena, interpret results, and even coordinate end-to-end research workflows. Supporters argue that FMs are reshaping the structure of scientific discovery by lowering entry barriers, facilitating exploratory iteration, and redistributing agency between humans and machines~\cite{bishop2022ai4science,elsevier2024ai}. Skeptics, however, view FMs as powerful yet conventional tools that amplify existing methodologies~\cite{wolfram2024ai,mcquillan2022resisting}. From this perspective, FMs serve to accelerate scientific progress without fundamentally transforming its underlying paradigm.

\begin{figure}[!t]
    \centering
    \includegraphics[width=\linewidth]{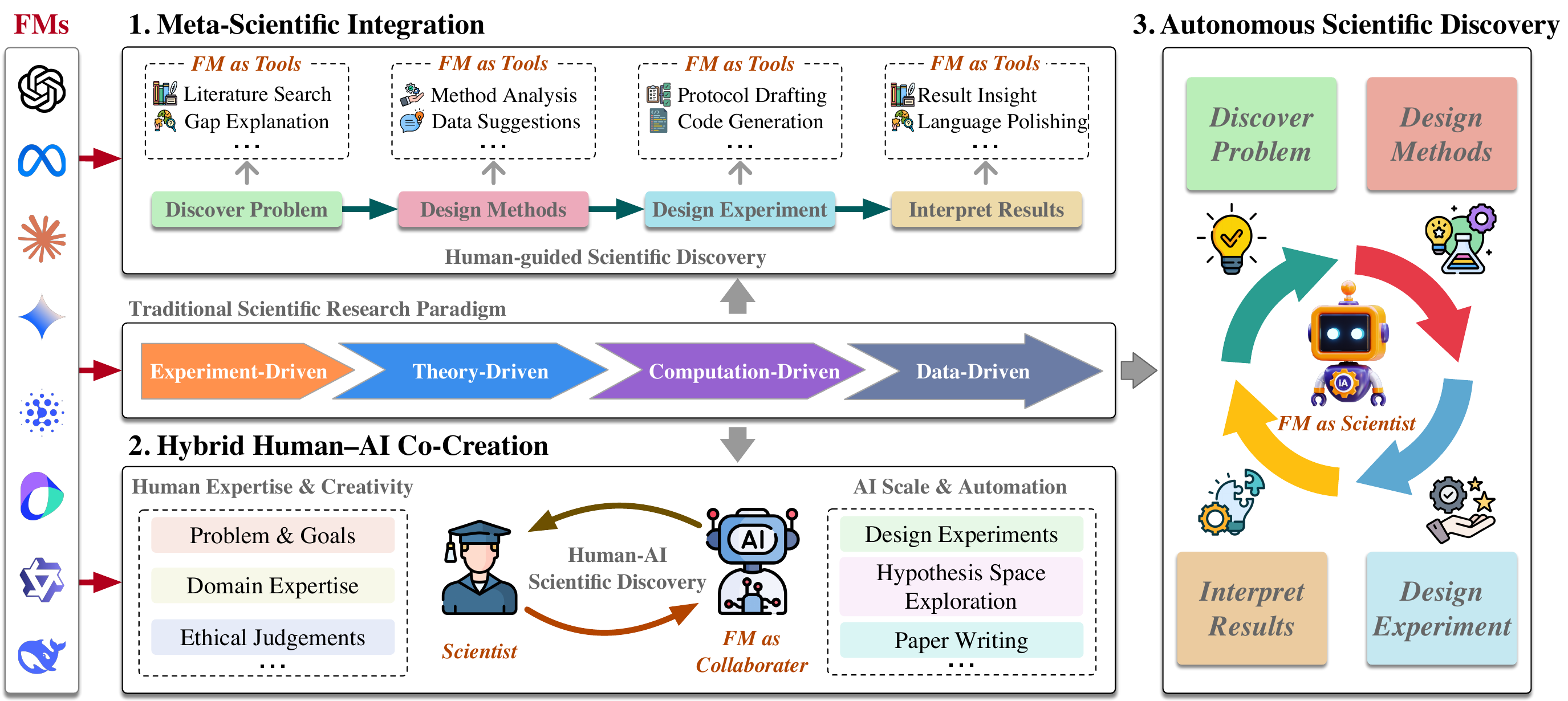}
    \caption{Evolving scientific paradigms empowered by FMs. FMs progressively transition from tool-like infrastructure (meta-scientific integration), to interactive co-creators (hybrid human–AI collaboration), and ultimately to autonomous agents capable of end-to-end scientific discovery.}
    \label{fig:scientific_discovery_paradigm}
\vspace{-0.2in}
\end{figure}

This paper enters that debate with a clear position: \textbf{FMs are not only improving parts of the scientific process, they are beginning to reshape the paradigm through which science operates.} To support this argument, we propose a three-stage framework that describes the progressive integration of FMs into scientific discovery, as illustrated in Figure~\ref{fig:scientific_discovery_paradigm}. (1) \emph{Meta-Scientific Integration}. FMs operate as flexible infrastructure across traditional scientific paradigms. They integrate cross-domain data, automate reasoning steps, and support end-to-end workflows, while remaining embedded within established epistemic structures. (2) \emph{Hybrid Human-AI Co-Creation}. In this transitional phase, FMs shift from passive tools to active collaborators. They participate in problem formulation, hypothesis generation, and experimental design, enabling more dynamic, iterative, and co-creative modes of discovery. (3) \emph{Autonomous Scientific Discovery}. Looking ahead, we envision FMs acting as autonomous agents of science. These systems will be capable of initiating questions, executing simulations, interpreting results, and generating new knowledge across both virtual and physical domains. At this stage, the scientific process becomes partially self-directed, presenting a shift toward a fundamentally new epistemic regime.

\textbf{Our Contributions}: (1) \emph{A new conceptual framework.} We introduce a three-stage framework to position FMs as catalysts of scientific paradigm evolution, spanning infrastructure support, collaborative reasoning, and autonomous discovery. (2) \emph{A systematic review and taxonomy.} We present a systematic analysis of FM-enabled scientific discovery, organized by their integration into experimental, theoretical, computational, and data-driven workflows. (3) \emph{A research agenda.} We identify key risks that should be addressed to realize the full scientific potential of FMs, and we also outline directions for future research on aligning epistemic goals with emerging AI capabilities.

\section{Background and Preliminary}
\vspace{-0.1cm}
\begin{wrapfigure}[13]{R}{0.6\textwidth}
  \vspace{-0.4cm}
  \centering
  \includegraphics[width=\linewidth]{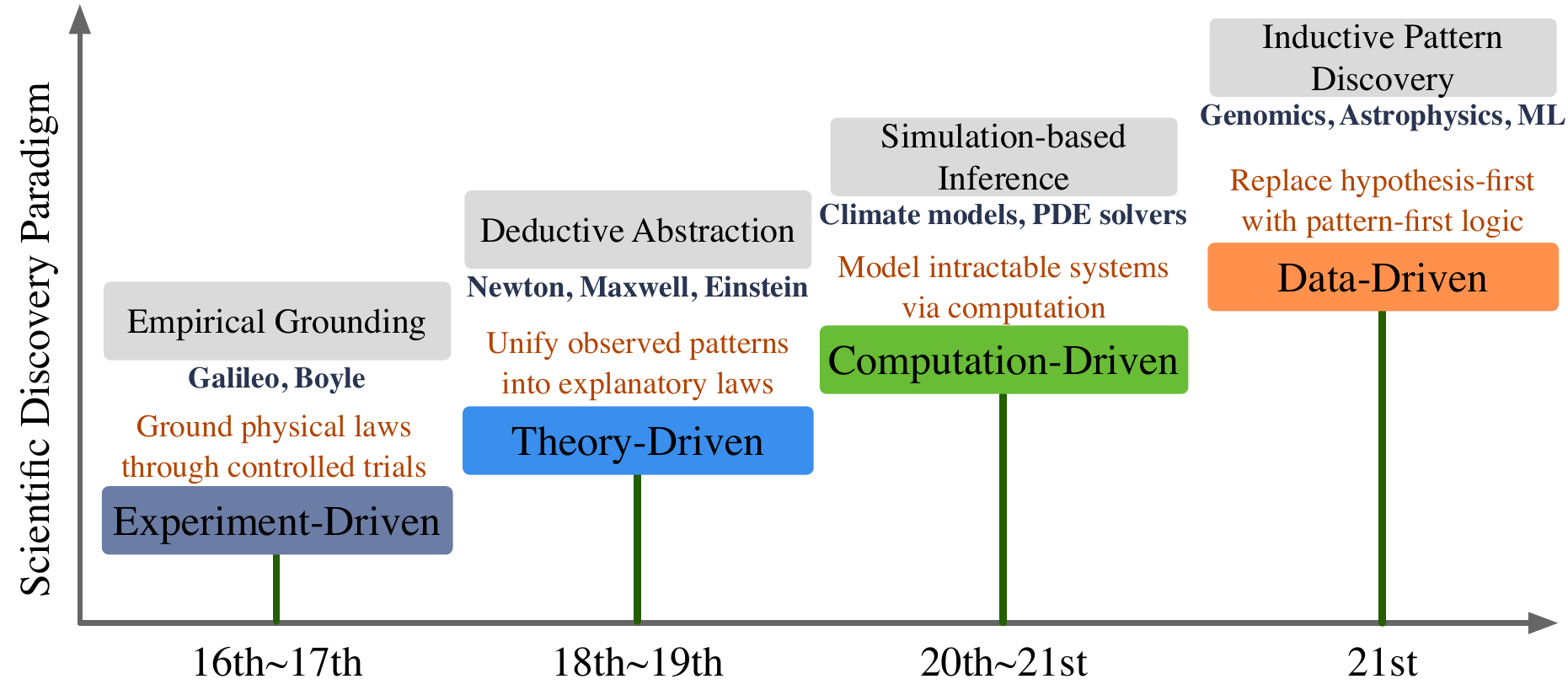}
    \caption{A roadmap of scientific discovery paradigms and their epistemic capabilities.}
    \label{fig:scientific_research_history}
\end{wrapfigure}

\textbf{A Brief History of Scientific Discovery.} Scientific discovery has advanced through a series of methodological paradigms, each reshaping the ways we observe, explain, and intervene in the natural world, as illustrated in Figure~\ref{fig:scientific_research_history}. These paradigms emerged in response to the limitations of their predecessors and were enabled by conceptual breakthroughs or technological innovations. Specifically, today’s scientific practice is shaped by four foundational paradigms: experiment-driven, theory-driven, computation-driven, and data-driven science. Each has introduced new standards for reasoning, validation, and knowledge generation. 

\begin{itemize}
    \item (1) The \emph{experiment-driven paradigm} arose during the scientific revolution of the 16\textsuperscript{th} and 17\textsuperscript{th} century, emphasizing systematic observation and controlled experimentation~\cite{shapin2018scientific}. Pioneers such as Galileo and Boyle designed repeatable experiments to validate natural laws~\cite{galilei1914two,boyle2013sceptical}, establishing measurability, verifiability, and reproducibility as empirical norms. However, this approach struggled with large-scale, highly complex, or inaccessible systems, where direct manipulation was impractical or impossible.
    \item  (2) The \emph{theory-driven paradigm} emerged in the 18\textsuperscript{th} and 19\textsuperscript{th} centuries, driven by advances in mathematics and formal logic~\cite{weyl2015symmetry}. Newton, Maxwell, and Einstein proposed abstract, unified theories that explained a wide range of phenomena under compact formulations~\cite{wigner1990unreasonable,einstein1922general}. While these models offered significantly expanded explanatory power, they also introduced a growing gap between theoretical complexity and empirical testability.
    \item  (3) The \emph{computation-driven paradigm} gained traction in the mid-20\textsuperscript{th} century with the advent of digital computing and numerical simulation~\cite{winsberg2019science,oreskes1994verification}. It enabled scientists to model systems that were either analytically intractable or experimentally inaccessible, such as global climate or molecular interactions. This gave rise to new forms of scientific reasoning, including scenario testing, model-based inference, and virtual experimentation. More recently, hybrid methods like physics-informed machine learning have further blurred the boundaries between theory, simulation, and data~\cite{karniadakis2021physics}.
    \item   (4) The \emph{data-driven paradigm} became prevalent in the 21\textsuperscript{st} century, fueled by exponential growth in sensing technologies, digitization, and computational power. This paradigm focuses on discovering patterns in high-dimensional data using statistical and machine learning techniques~\cite{breiman2001statistical,lecun2015deep}. Applications span diverse fields from genomics to astrophysics~\cite{segal2003module,ivezic2019lsst}, enabling data-driven insights in domains where prior models were lacking or underdeveloped. Despite its success, data-centric science often struggles with causal inference, interpretability, and robustness under distributional shifts~\cite{rudin2019stop,pearl2009causality,scholkopf2021toward}.

\end{itemize}

The above four paradigms have progressively expanded the scope and scale of scientific discovery. However, as scientific problems become increasingly complex and interdisciplinary, the limitations of each paradigm, particularly when applied in isolation, become more apparent. This calls for rethinking how scientific discovery might evolve beyond the current frameworks.

\textbf{Foundation Models.}
FMs are large-scale neural networks trained on massive and diverse datasets, designed to serve as general-purpose systems adaptable to a wide range of downstream tasks~\cite{DBLP:journals/corr/abs-2303-08774, DBLP:journals/corr/abs-2412-19437, Grok_3,  radford2021learning, qwen2.5, DBLP:journals/corr/abs-2410-21276, liu2025bag}. They are typically developed through unsupervised or self-supervised pertaining, for example, by predicting masked tokens in text or aligning image–captions pairs, followed by task-specific fine-tuning or prompting. This pretrain and fine-tune paradigm allows knowledge acquired during large-scale training to transfer across domains and tasks. The emergence of FMs marks a shift from narrow, task-specific models to flexible systems capable of generalizing across modalities and problem types. This is particularly valuable in scientific domains, where labeled data is scarce, tasks are often open-ended, and disciplinary boundaries are increasingly fluid. FMs provide a unified modeling framework that integrates language, vision, code, and structured data, enabling diverse tasks in reasoning, generation, and retrieval.
Prominent examples include GPT-4~\cite{, DBLP:journals/corr/abs-2410-21276}, a language model that performs a wide range of tasks, including question answering, summarization, and code generation, through zero-shot or few-shot prompting. Another is CLIP~\cite{radford2021learning}, a vision–language model trained on 400 million image–text pairs using a contrastive learning objective. Without additional fine-tuning, CLIP can classify images based on natural language prompts, demonstrating strong zero-shot capabilities. More recently, FMs have expanded into scientific domains such as protein folding (e.g., AlphaFold~\cite{jumper2021highly}), mathematical discovery (e.g., FunSearch~\cite{romera2024mathematical}), and robotics~\cite{Yoshikawa2023}. These applications highlight the growing role of FMs not merely as tools for automation but as general-purpose engines for scientific reasoning, interpretation, and discovery.

\section{Rethinking Scientific Paradigms in the Era of Foundation Models}~\label{sec:Scientific_Paradigms_FM}
In this section, we introduce a three-stage framework to describe the evolving role of FMs in scientific discovery and to explore their potential for paradigm-level change based on the degree of autonomy, task scope, as illustrated in Table~\ref{tab:FM_Paradigms}. We argue that FMs are not only enhancing individual components of the scientific process but are also beginning to reshape the broader structure of scientific discovery. Although this transformation is still in its early stages, the shift is already underway.

\begin{table}[ht]
\vspace{-0.1cm}
\centering
\caption{Comparison of FM Paradigms Across Five Dimensions}~\label{tab:FM_Paradigms}
\scalebox{0.82}{
\begin{tabular}{>{\raggedright\arraybackslash}p{3.2cm}
                >{\centering\arraybackslash}p{4cm}
                >{\centering\arraybackslash}p{4cm}
                >{\centering\arraybackslash}p{4cm}}
\hline
\textbf{Dimension} & \textbf{Meta-Scientific Integration} & \textbf{Hybrid Human--AI Co-Creation} & \textbf{Autonomous Scientific Discovery} \\
\hline
\textbf{Paradigm Definition} & Tool   & Human–AI collaborator & Independent agent \\
\textbf{FM Role} & Backend tool & Co-creator & Autonomous actor \\
\textbf{Task Scope} & Task enhancer & Full-cycle tasks & End-to-end, self-directed \\
\textbf{Autonomy} & Low & Moderate & High \\
\textbf{Impact on Science} & Efficiency boost & Labor shift & Scientific re-foundation \\
\hline
\end{tabular}
}
\end{table}

\textbf{Meta-Scientific Integration.} In the meta-scientific integration mechanism,  FMs function as an intelligent infrastructure that augments, but does not transform, scientific practice. Their core role lies in streamlining fragmented processes, enhancing interoperability, and increasing the reproducibility and efficiency of workflows across disciplinary boundaries. FMs in this paradigm serve as backend coordinators: they automate procedural tasks such as data preprocessing, literature retrieval, and methodology matching. By integrating components that were once isolated (e.g., linking sensor data with simulation models or connecting experimental planning with prior knowledge), FMs facilitate smoother, modular research pipelines. However, these systems remain fully bound by human-defined objectives and lack the capacity to initiate or reframe scientific inquiry. Crucially, the role of FMs here is instrumental, not epistemic. They execute tasks within established scientific paradigms without altering their logic or structure. Despite their technical sophistication, they exhibit low autonomy and require continuous human supervision. This paradigm is therefore augmentative rather than transformative: it improves how science is conducted but does not redefine what science fundamentally is. FMs increase scientific throughput and integration, but the locus of reasoning and knowledge production remains firmly human.

\textbf{Hybrid Human-AI Co-Creation.}
FMs are shifting from passive infrastructure to active collaborators within scientific workflows. Rather than remaining in the background, they now work alongside human researchers in shared reasoning and decision-making processes, enabling a hybrid intelligence model that pairs human intuition and expertise with the generalization, memory, and automation capabilities of FMs. In this new role, FMs assist across the scientific pipeline, contributing to research question generation, hypothesis structuring, experiment planning, and, in some cases, end-to-end task execution. Their involvement extends beyond operational support to ideation and interpretation, though they continue to operate within boundaries set by human oversight and scientific norms. The scope of FM contributions has expanded significantly. They engage in hypothesis generation, problem scoping, experiment design, execution support through automation, interpretation of findings, and participation in scientific discourse. While their functions span the full research cycle, their actions are still initiated, constrained, and validated by humans. FMs exhibit moderate autonomy: they can generate ideas, select methods, and adapt workflows based on feedback within scoped research environments, but they rely on human prompts for problem framing and ethical guidance. Their outputs can influence the trajectory of discovery but do not fully determine it. This evolving paradigm begins to reshape the division of cognitive labor in science. By offloading tasks such as literature synthesis, multi-step reasoning, and combinatorial experiment planning, FMs allow human researchers to focus more on judgment, creativity, and strategic framing. Although the human-AI co-creation model introduces a new form of epistemic collaboration, it does not yet constitute standalone scientific intelligence. FMs reconfigure how science is conducted without redefining who conducts it.

\textbf{Autonomous Scientific Discovery.}
In this emerging paradigm, FMs take a decisive step beyond collaboration, evolving into autonomous agents capable of conducting scientific discovery with minimal or no human oversight. Unlike earlier stages, where FMs assist under predefined goals or structured prompts, autonomous FMs can initiate and carry out the entire scientific cycle on their own, i.e., posing research questions, generating hypotheses, selecting methods, executing experiments or simulations, interpreting results, and updating internal models based on outcomes.
What distinguishes this paradigm is the high degree of autonomy and continuity in the reasoning process. FMs are no longer reactive tools triggered by human input. They operate in a self-directed manner, guided by internal objectives and feedback mechanisms. Their behaviors resemble those of scientific investigators: identifying promising research directions, exploring solution spaces, evaluating novelty and coherence, and refining strategies based on intermediate findings.
Such agentic behavior enables FMs to function not just as tools, but as epistemic actors that can contribute original insights, challenge existing theories, and shape the direction of scientific discourse. These models possess the capacity to synthesize diverse knowledge sources, bridge conceptual gaps across disciplines, and adapt dynamically to new evidence or goals. They are capable of making decisions about what to explore, how to explore it, and when to revise their understanding, all without explicit instruction.
The broader implications are profound. If fully realized, this paradigm would mark a fundamental shift in the structure of scientific discovery. Rather than simply accelerating human-led research, FMs would become independent engines of scientific discovery, redefining who or what can produce scientific knowledge. This shift introduces what we call the fifth scientific paradigm, where discovery is no longer exclusively human-driven but emerges from the autonomous reasoning of machine intelligence. While still in its formative stages, this trajectory is already taking shape in systems like AI scientists~\cite{ai_scientist}, which have conducted the whole research pipeline to solve scientific problems. As FMs continue to evolve, their role may expand beyond assistance or collaboration toward initiating, directing, and validating new lines of scientific investigation. This transformation challenges long-held assumptions about the nature of scientific agency and opens new questions about responsibility, trust, and validation in machine-led discovery.
\vspace{-0.1cm}
\section{Foundation Model Integration Across Scientific Paradigms}
\vspace{-0.1cm}
\eat{Building on the three-part framework introduced in Section~\ref{sec:Scientific_Paradigms_FM}, this section examines how foundation models (FMs) are instantiated across the four classical scientific paradigms: experimental, theoretical, computational, and data-driven science. We review recent advances demonstrating how FMs are increasingly integrated into scientific methodologies, for example, automating sub-tasks in experimental workflows, supporting symbolic reasoning in theoretical modeling, and identifying patterns or generating hypotheses in large-scale data environments. Looking ahead, we envision the emergence of a new paradigm of autonomous scientific discovery, in which FMs unify observation, modeling, and simulation within an end-to-end, self-directed research pipeline. \TODO{The taxonomy of foundation model integration across scientific paradigms is shown in Figure~x.}}

Building on our three-stage framework of FM-driven scientific evolution, this section spans the first two stages, Meta-Scientific Integration and the early signs of Hybrid Human-AI Co-Creation, by examining how FMs are increasingly embedded within and across classical scientific paradigms. We systematically review their roles in experimental, theoretical, computational, and data-driven scientific discovery, as well as their emerging capacity to mediate cross-paradigm workflows.

\subsection{Experiment-Driven Paradigm}

The experiment-driven paradigm emphasizes empirical observation, controlled intervention, and iterative refinement. However, traditional workflows are constrained by limited planning capacity, costly trial spaces, and brittle automation pipelines. FMs offer new opportunities to enhance this paradigm by improving design efficiency and enabling more flexible, adaptive execution. Current integrations focus on two key stages: (1) experiment design and (2) physical experiment execution.

\textbf{Experimental Design.}
Designing informative experiments under resource constraints remains a core scientific challenge. Classical methods such as Bayesian optimization (BO) and active learning (AL) often suffer from sparse priors and poor generalization. FMs help overcome these issues by encoding domain knowledge and guiding the search for optimal configurations~\cite{roohani2025biodiscoveryagentaiagentdesigning, nguyen2023exptsyntheticpretrainingfewshot, ma2024optogptfoundationmodelinverse, jia2024llmatdesignautonomousmaterialsdiscovery}. For instance, FMs serve as priors or feature extractors in BO pipelines, accelerating convergence in molecular and materials discovery \cite{kristiadi2024soberlookllmsmaterial, cissé2025languagebasedbayesianoptimizationresearch}. Building on this,  FMs further improve data efficiency by directly maximizing mutual information, bypassing the need for surrogate modeling \cite{iollo2025bayesianexperimentaldesigncontrastive}. These approaches point toward a future in which FMs co-adapt with experimental design processes, forming the backbone of closed-loop, context-aware optimization agents. 

\textbf{Physical Experiment Execution.}
In laboratory settings, executing experiments demands coordination across planning, perception, and control domains that are traditionally fragmented and manually programmed. FMs increasingly act as unifying interfaces and planners~\cite{Boiko2023, Ruan2024, Yoshikawa2023, F_bba_2025, mathur2024visionmodularaiassistant}. For example, FMs have been employed to generate Python control scripts for scientific instruments, translating user-specified objectives into directly executable lab protocols \cite{F_bba_2025}, while LLM-RDF orchestrates modular agents for structured reaction planning \cite{Ruan2024}. More dynamic systems, such as CLAIRify, embed FMs into robotic control, enabling physical manipulation through language-guided planning \cite{Yoshikawa2023}. Pushing further, multimodal agents like VISION and AP-VLM incorporate vision and speech to support real-time interaction and error correction in lab environments \cite{mathur2024visionmodularaiassistant, sripada2024apvlmactiveperceptionenabled}.

\subsection{Theory-Driven Scientific Paradigm}
The theory-driven paradigm seeks to construct formal, generalizable frameworks that explain observed phenomena and yield testable predictions. Traditionally dependent on human intuition and symbolic logic, this paradigm has been constrained by limited idea generation, steep formalization requirements, and the brittleness of proof systems. FMs are increasingly being integrated to augment this process by expanding the space of plausible hypotheses and supporting formal reasoning pipelines that validate theoretical claims. 

\textbf{Scientific Hypothesis Generation.}
FMs facilitate systematic hypothesis generation by synthesizing knowledge across large-scale corpora and structured priors~\cite{alkan2025surveyhypothesisgenerationscientific, guo2024embracing, oneill2025sparkssciencehypothesisgeneration, chai-etal-2024-exploring}. Rather than relying solely on intuition or fragmented evidence, recent methods guide FMs using knowledge graphs and domain constraints. For example, KG-CoI steers hypothesis formulation through ontological concept paths to enhance novelty and verifiability \cite{guo2024embracing}. The HypoGen dataset further improves model outputs by grounding them in historical patterns of scientific idea evolution, improving both creativity and feasibility \cite{oneill2025sparkssciencehypothesisgeneration}. In scientific domains like physics and climate modeling, physics-guided foundation models embed physical laws directly into the generation process to ensure consistency with known dynamics \cite{farhadloo2025physicsguidedfoundationmodels, yu2025physicsguidedfoundationmodelscientific}.

\textbf{Theory Validation and Formal Reasoning.}
To validate hypotheses, FMs are increasingly linked with symbolic logic systems to support deductive inference, consistency checking, and falsifiability analysis~\cite{pan2023logiclmempoweringlargelanguage, dinu2024symbolicaiframeworklogicbasedapproaches, cunnington2024rolefoundationmodelsneurosymbolic, huang2025automatedhypothesisvalidationagentic, song2025leancopilotlargelanguage, xin2024deepseekproveradvancingtheoremproving}. Logic-LM exemplifies this by coupling LLMs with symbolic solvers in a feedback loop, improving formal rigor in logical tasks \cite{pan2023logiclmempoweringlargelanguage}. General-purpose neuro-symbolic systems like SymbolicAI and Vieira extend this framework to domain-specific reasoning tasks \cite{dinu2024symbolicaiframeworklogicbasedapproaches, Li_2024}. For automated theory testing, the Popper system uses LLMs to generate counterexamples and identify falsifiable conditions \cite{huang2025automatedhypothesisvalidationagentic}. In formal mathematics, LeanCopilot and DeepSeekProver demonstrate the capacity of pretrained models to assist in proof construction and verification at scale \cite{song2025leancopilotlargelanguage, xin2024deepseekproveradvancingtheoremproving}.

\subsection{Computation-Driven Scientific Paradigm}

The computation-driven paradigm advances scientific discovery through the formulation and execution of mathematical models that simulate, predict, or control complex systems. While traditional workflows depend on hand-crafted equations and high-cost numerical solvers, they often face limitations in flexibility, scalability, and automation.  FMs offer new capabilities by enabling automated model construction and accelerating scientific computation. We review recent progress along two key fronts: constructing executable scientific models and efficiently solving or inverting them.

\textbf{ Formulating Executable Scientific Models.} Conventional scientific modeling relies on expert-designed equations or symbolic regressors, which struggle with multi-scale dynamics, noisy observations, and sparse priors. FMs enhance this process by supporting symbolic, latent, and differentiable formulations. For instance, in symbolic discovery, systems like \textsc{LLM-SR} translate diverse inputs such as plots or text into equation skeletons for subsequent refinement, while others like \textsc{FunSearch} discover new algorithms by framing program synthesis as a language-guided search task \cite{shojaee2024llm, DeepMind2023Funsearch, Grayeli2024LaSR}. When explicit equations are elusive, FMs excel at learning latent operators. \textsc{PROSE-PDE}, for example, simultaneously predicts system dynamics and infers underlying governing laws within a learned representation, and \textsc{DiffusionPDE} trains generative priors over coefficient-solution pairs to sample posteriors from sparse data, effectively bypassing direct equation formulation \cite{Liu2024PROSEPDE, Huang2023DiffusionPDE}. Such modeling methods unify forward and inverse modeling under shared representations.

\textbf{Solving and Inverting Scientific Equations.} Once scientific models are formulated, whether as partial differential equations or latent operator representations, solving and inverting them remains computationally demanding. Classical methods typically require spatial discretization, expert-crafted solvers, and often fragile optimization routines, especially in ill-posed or high-dimensional settings. FMs operate directly over function spaces, guided by learned priors and generative inference mechanisms, to accelerate solutions and enable efficient inversion. A pivotal development on Neural Operator learns continuous maps from forcing terms to partial differential equation (PDE) solutions, generalizing across mesh resolutions and becoming a cornerstone for physics surrogates \cite{Kovachki2021NeuralOperator, Rahman2024CoDANO}. Building on this, models like \textsc{GraphCast} now outperform traditional numerical weather prediction models at reduced computational cost, and specialized architectures like \textsc{FactFormer} handle massive computational grids \cite{Lam2023GraphCast, Li2023FactFormer}. Furthermore, FMs can enhance legacy solvers; for example, PDE-Refiner architectures iteratively correct coarse solver output, trimming error without rerunning the full simulation \cite{lippe2023pde}. These innovations significantly reduce the computational burden.

\subsection{Data-Driven Scientific Paradigm}

The data-driven paradigm begins with large-scale observations collected across instruments, populations, and modalities. It aims to discover latent scientific structures and generate predictive outputs directly from data, often without recourse to explicit physical models. Traditional workflows rely on handcrafted features, narrow supervision, and unimodal pipelines, limiting their ability to scale, integrate, or generalize. FMs offer a unified upgrade by learning statistical regularities across domains and enabling flexible reasoning over heterogeneous signals. Current applications cluster into two major directions: (1) scientific knowledge discovery from multimodal data and (2) predictive scientific inference through generative modeling.

\textbf{ Scientific Knowledge Discovery.} Classical methods for extracting scientific knowledge, such as enrichment analysis or rule mining, struggle with noisy, multimodal, or unstructured data. FMs address these limitations by compressing vast corpora into structured representations and supporting inference across modalities. For instance, token-based FMs like \textsc{DNABERT} identify functional DNA elements from sequences \cite{ji2021dnabert}. In chemistry, \textsc{MoLFormer} learns SMILES embeddings that correlate linearly with key molecular properties, enabling zero-shot retrieval of candidate molecules \cite{molformer2022}. Beyond single modalities, multimodal FMs like \textsc{ChemVLM} integrates molecular structure images and textual descriptions to answer complex multimodal chemistry questions \cite{Li2024ChemVLM, chemcrow}. In the spatio-temporal domain, \textsc{ClimaX}\cite{nguyen2023climax} fuses diverse climate inputs, spanning reanalysis data, climate model simulations, and satellite observations, learning unified spatio-temporal representations through masked autoencoding. These rich embeddings capture underlying climate patterns and their complex interdependencies, thereby facilitating the discovery of novel insights into Earth system dynamics and the characterization of various climate phenomena. Furthermore, large language models pretrained on vast corpuses of scientific literature, such as \textsc{Galactica}, act as powerful tools to organize, synthesize, and query scientific knowledge, effectively transforming millions of papers into an accessible and computationally tractable knowledge base \cite{Taylor2022Galactica, medpalm2}.

\textbf{ Predictive Scientific Inference.} In many domains, predictive accuracy is now more critical than explicit mechanistic modeling. Classical surrogate models, however, often struggle with high-dimensionality and uncertainty. FMs redefine this task as generative modeling, trained directly on observational or simulation-derived data. For instance, in spatiotemporal forecasting, \textsc{GraphCast} and \textsc{Pangu-Weather} learns latent dynamics from re-analysis datasets to produce global weather predictions rivaling numerical models at lower computational cost \cite{sanchez2024graphcast, Bi2023Pangu}. Diffusion-based models like \textsc{DiffusionSat} can generate high-resolution satellite imagery from coarser inputs, bridging observational gaps \cite{Khanna2024DiffusionSat}. In structural prediction, FMs such as \textsc{AlphaFold 2} and \textsc{ESMFold} predict protein structures from sequences with near-experimental accuracy \cite{jumper2021alphafold, esmfold}. Furthermore, generative models like \textsc{RFdiffusion} can design novel protein folds and functional interfaces, while \textsc{MatterGen} extends the same paradigm to inorganic crystal design, producing stable materials that satisfy user-specified property constraints \cite{watson2023rfdiffusion,zeni2025generative}, demonstrating the capacity of FMs to turn data into actionable foresight.

\subsection{Cross-Paradigm Foundation Model Integration}

Classical scientific paradigms, experimental, theoretical, computational, and data-driven, have historically represented distinct methodological lenses, though they are often employed in combination in scientific practice. As modern scientific challenges grow in complexity, discovery increasingly relies on workflows that integrate these paradigms into unified, cross-cutting pipelines. FMs, with their general-purpose reasoning abilities, multimodal interfaces, and growing autonomy, are uniquely positioned to mediate such integrative, hybrid workflows.

Recent advances show that FMs can serve as integrative engines across classical scientific paradigms, experimental, theoretical, computational, and data-driven, by enabling workflows that traverse and connect traditionally siloed approaches. Crucially, these models maintain interpretability and cross-domain transferability, supporting scientific reasoning that is both coherent and generalizable across diverse methodologies~\cite{yamada2025ai, agent_lab, AI_ResearchAgent, DBLP:journals/corr/abs-2408-06292, liu2025mm}. For example, PROSE-FD \cite{Sun2024PROSE} co-trains symbolic equation templates and spatial field data within a multimodal Transformer, enabling cross-regime generalization in fluid dynamics and jointly discovering both structure and solution behavior. Similarly, Latent Neural Operators (LNOs) \cite{Wang2024LNO} encode physical operators into latent spaces that are geometry-agnostic and resolution-invariant, allowing both forward and inverse problems to be solved within a shared learned representation. Beyond individual modeling components, FMs increasingly orchestrate end-to-end scientific workflows that couple theory, simulation, data, and experimentation. In chemistry, for instance, systems like Coscientist \cite{Boiko2023} translate high-level research goals into machine-executable protocols, control robotic synthesis, and adapt future actions based on experimental results.

\section{Risks of Emerging FM-Centered Scientific Paradigms and Future Direction}
In this section, we introduce the key risks posed by emerging FM-centered scientific paradigms, including challenges related to bias, misinformation, reproducibility, and scientific accountability.  We then outline future directions toward autonomous scientific discovery, highlighting embodied agents, closed-loop workflows, and continual learning.

\subsection{Risks of Emerging FM-driven Scientific Paradigms}

While FMs promise transformative benefits across the scientific enterprise, their growing autonomy introduces critical epistemic, technical, and ethical risks. These risks evolve and intensify as FMs transition from backend tools (Meta-Scientific Integration) to collaborative partners (Hybrid Human–AI Co-Creation), and ultimately toward independent research agents (Autonomous Scientific Discovery). We identify four key risk dimensions that require anticipatory mitigation to ensure the responsible evolution of FM-driven scientific paradigms.

\textbf{Bias and Epistemic Fairness.}
Even in early-stage applications, such as literature review and task assistance, FMs inherit biases from their training data, which often overrepresent dominant paradigms, Western institutions, and widely cited authors~\cite{wang2023aleatoric, cortes2023statistical}. As FMs transition into co-creators and autonomous agents, these biases shift from being passive reflections to active forces shaping scientific agendas. For instance, in global health modeling, a FM trained predominantly on English-language publications and high-impact journals may systematically prioritize research on diseases like Type 2 diabetes or cardiovascular conditions, topics well-studied in Western contexts, while overlooking pressing but underrepresented issues such as schistosomiasis or child stunting in sub-Saharan Africa~\cite{gan2024erasing, gan2024erasing}.  Without intervention, this can lead to epistemic homogenization and the exclusion of underrepresented perspectives. Mitigating these risks calls for more diverse and inclusive training datasets, targeted fine-tuning on marginalized knowledge domains, and fairness-aware evaluation protocols embedded throughout the FM pipeline.

\textbf{ Hallucination and Scientific Misinformation.}
Across all paradigms, FMs remain fundamentally data-driven pattern recognizers rather than truth-preserving reasoners. As their role shifts from task augmentation to autonomous hypothesis generation, the risk of generating plausible but unverified claims grows substantially~\cite{dhuliawala2023chain,wu2024iter}. In biomedical domains, for instance, an FM might propose a novel mechanism that appears convincing but lacks experimental grounding, potentially misguiding research efforts. In physics, it may generate elegant but physically invalid formulations. These failures can propagate if outputs are prematurely trusted or cited. To mitigate this, FMs should incorporate verification mechanisms such as symbolic logic checks, simulation-based validation, human-in-the-loop review, and provenance tracking to ensure traceability and scientific credibility.

\textbf{ Reproducibility and Scientific Transparency.}
As FMs take on more end-to-end responsibilities, such as designing experiments, running simulations, and interpreting results, their decision-making processes often remain opaque
~\cite{mehta2024openelm, wolter2025position}.This threatens scientific reproducibility: without visibility into intermediate reasoning steps, model assumptions, or version states, it becomes difficult to replicate or validate outcomes. For example, a model-generated chemical synthesis pathway may lack interpretable derivations. Addressing this requires transparent logging of reasoning steps, version-controlled model checkpoints, and open-science practices that preserve the traceability of FM-driven scientific workflows.

\textbf{Authorship, Accountability, and Scientific Ethics.}
As FMs shift from tools to collaborators and, ultimately, autonomous agents, questions around intellectual credit, accountability, and ethical conduct become increasingly urgent~\cite{chakraborty2025hallucination}. If an FM generates a core hypothesis or experimental design, should it be acknowledged as a co-author? Who is accountable if its output causes harm or leads to flawed science? While such issues were peripheral in earlier paradigms, they become central in autonomous discovery. Risks include ghost authorship, diminished human contribution, and misuse of FM-generated content. Addressing these concerns requires governance frameworks that distinguish mechanical from creative contributions, mandate transparent disclosures, and track downstream impacts of AI-generated outputs.

\subsection{Future Directions: Toward Autonomous Scientific Discovery}
\vspace{-0.1in}
Despite recent advances, most FM deployments remain confined to static prompts, predefined tasks, and fixed schema. They typically lack persistent memory, adaptive feedback, and physical embodiment. As such, their contributions, though impressive, are largely reactive and limited to isolated stages of the scientific process. Looking ahead, we identify three concrete research directions that define the transition toward autonomous scientific discovery:

\textbf{Embodied Scientific Agents.}
A pivotal step toward scientific autonomy is grounding FMs in the physical world. Future FMs will increasingly be deployed within laboratory robotics, automated instruments, and digital twin environments. By coupling language-based reasoning with real-world perception and control, these agents will plan experiments, interact with physical systems, and iteratively refine procedures. This integration of abstract reasoning with physical execution is essential for closing the loop between scientific modeling and empirical verification. However, challenges remain in integrating high-level task planning with low-level control, ensuring robustness under real-world uncertainty, and maintaining safety and interpretability in dynamic lab environments. 

\textbf{Closed-Loop Scientific Autonomy.}
Current scientific workflows are typically open-loop: FMs assist with parts of the pipeline, but humans still decide the next steps. Moving toward truly autonomous science requires closed-loop systems, where FMs continuously formulate hypotheses, design and perform experiments, analyze results, and update internal models based on feedback. Current progress includes reinforcement learning-based planning~\cite{weng2024cycleresearcher}, planning-as-inference~\cite{ren2025towards, botvinick2012planning}, and neuro-symbolic agents~\cite{li2025proving}.  For example, recent neuro-symbolic agents have shown how structured memory and logic-based reasoning can guide molecule design or theorem proving~\cite{shojaee2024llm}. Similarly, planning-as-inference approaches~\cite{qiao2024agent} and reinforcement learning-based agents~\cite{bader2022leveraging} have been applied to automate scientific workflows such as hypothesis selection and experimental sequencing. A key challenge is ensuring that the loop remains robust to noisy observations, adaptive to shifting objectives, and aligned with scientific validity, not just reward maximization.  

\textbf{Continual Learning and Generalization.}
To operate effectively across scientific domains, FMs must transition from static systems to continual learners capable of accumulating and refining knowledge over time. This entails addressing key challenges such as catastrophic forgetting~\cite{ostapenko2022continual} and domain drift~\cite{chen2024dual}. Promising approaches include parameter-efficient online adaptation~\cite{nuertey2025parameter}, memory-augmented architectures~\cite{wang2023augmenting}, and modular lifelong learning frameworks~\cite{yu2024boosting} that allow selective knowledge retention and update. However, existing methods still fall short in enabling robust transfer across heterogeneous tasks and modalities. Advancing continual learning mechanisms would allow FMs to incrementally build domain-bridging representations, facilitate analogical reasoning across scientific contexts, and sustain coherent research trajectories over extended periods~\cite{yang2025recent}.

\vspace{-0.1cm}
\section{Conclusions}
\vspace{-0.1cm}
FMs are reshaping the landscape of scientific discovery. From enhancing existing workflows to enabling autonomous inquiry, they signal a potential shift toward a fifth scientific paradigm. In this paper, we proposed a three-stage framework, \ie meta-scientific integration, hybrid human-AI co-creation, and autonomous scientific discovery, to characterize this evolving trajectory. By analyzing FM integration across classical paradigms, we showed how FMs increasingly act not only as tools but as epistemic agents. While this transformation is still emerging, it raises profound questions about agency, authorship, and the nature of knowledge itself. Looking forward, we call for rigorous exploration of FM capabilities, responsible governance mechanisms, and deeper theoretical understanding to guide their role in science. Embracing this shift may redefine not just how we do science, but who or what can do science.

{\small
\bibliographystyle{unsrt}
\bibliography{ref}
}

\end{document}